\UseRawInputEncoding 
\documentclass[a4paper,conference]{IEEEtran}
\usepackage{multirow} 
\usepackage[noadjust]{cite}
\ifCLASSINFOpdf
\usepackage[pdftex]{graphicx}
\graphicspath{{../pdf/}{../jpeg/}}
\DeclareGraphicsExtensions{.pdf,.jpeg,.png}
\else
\usepackage[dvips]{graphicx}
\graphicspath{{../eps/}}
\DeclareGraphicsExtensions{.eps}
\fi

\usepackage{amsmath}
\usepackage{algorithmic}
\usepackage{array}
\ifCLASSOPTIONcompsoc
\usepackage[caption=false,font=normalsize,labelfont=sf,textfont=sf]{subfig}
\else
\usepackage[caption=false,font=footnotesize]{subfig}
\fi

\usepackage{dblfloatfix}
\usepackage{url}
\begin{document}

		\title{Global Regular Network for Writer Identification}
		
		\author{\IEEEauthorblockN{Shiyu Wang}
			\IEEEauthorblockA{Xi’an Jiaotong University\\
				Xi’an, China\\
				Email: wangshiyu186@stu.xjtu.edu.cn}
			}
		
		\maketitle
		
		\begin{abstract}
Writer identification has practical applications for forgery detection and forensic science. Most models based on deep neural networks extract features from character image or sub-regions in character image, which ignoring features contained in page-region image. Our proposed global regular network (GRN) pays attention to these features. GRN network consists of two branches: one branch takes page handwriting as input to extract global features, and the other takes word handwriting as input to extract local features. Global features and local features merge in a global residual way to form overall features of the handwriting. The proposed GRN has two attributions: one is adding a branch to extract features contained in page; the other is using residual attention network to extract local feature. Experiments demonstrate the effectiveness of both strategies. On CVL dataset, our models achieve impressive 99.98\% top-1 accuracy and 100\% top-5 accuracy with shorter training time and fewer network parameters, which exceeded the state-of-the-art structure. The experiment shows the powerful ability of the network in the field of writer identification. The source code is available at \url{https://github.com/wangshiyu001/GRN}.
		\end{abstract}

		\IEEEpeerreviewmaketitle

		\section{Introduction}
		Just like language, fingerprints, iris and human faces, handwriting is an important feature of person. Writer identification based on handwriting plays a significant role in finance, insurance, public security departments, criminal investigations, court trials and other fields. However, handwriting is different from other biological characteristics. It mainly includes: (1) The handwriting of people at different ages will change; (2) At different time periods and locations, the writer will write changeable handwriting. How to recognize people's identity in changing handwriting is a huge challenge in the field of computer vision. The early method of writer identification based on feature point extraction. Many feature extraction algorithms are applied in this field\cite {HELLI20102199,CHRISTLEIN2017258,8270096,8395190,BERTOLINI20132069,YILMAZ2016109}. In recent years, Convolutional Neural Networks have achieved great success and were quickly applied in this field\cite {10.1007/978-3-319-23117-4_3,10.1007/978-3-319-24947-6_45,7814125,7814128,NGUYEN2019104,9040654}. However, most of the handwriting identification systems based on convolutional neural networks pay attention to local features extraction. But in fact, the characteristics of handwriting not only include local characteristics such as the thickness of the strokes and the degree of curvature, but also many global characteristics. In most cases, human handwriting is not written in a grid, but on empty white paper. The overall characteristics such as the arrangement of characters, and the arrangement of lines also affect the writer's inherent writing habits. For example, we often say that a person's writing is neat or scribbled, in which we focus on the global features of the handwriting rather than the local features. Therefore, the recognition of handwriting not only needs to pay attention to the local features, but also the overall features.
		
		The global regular network (GRN) pays attention to the global characteristics of the handwriting. The input of the network consist of two parts: the page handwriting and the word handwriting from the same writer: the page handwriting at the global level, which contains numerous global features; the word handwriting at the local level, which contains numerous local features. Our network extracts these two handwriting features in two branches. One is the global feature extraction network, which takes the page handwriting as input to extract the global features; the other is the local feature extraction network, which takes the word handwriting as the input to extract the local features. Finally, the global feature works on the local features in the way of parameter weights to form the overall feature of the handwriting. Our network specifically uses one network to extract the global features and combines them with local features to extract more abundant handwriting features.
		
		Besides a branch to extract global features, residual attention network proposed by Wang et al.\cite{8100166} is used to extract the local features of word handwriting. Features of word handwriting were always contained in special places like the corner of strokes and the attention mechanism\cite{WOS:000181471800047,Itti2001,WOS:000452647102103,7807286} used by residual attention network will help to extract these features. Sec.\ref{sec4-3} shows improved performance of GRN with residual attention network.
		
		The network structure has two novel highlights: On the one hand, different from the traditional focus on local feature extraction of handwriting, this network has adds a global feature extraction network to extract the global features of the handwriting and combine it with the local features to characterize the features of the handwriting. With this method, GRN can extract handwriting features at multiple levels. On the other hand, our network uses the residual attention network to extract the local features of the handwriting, which more improves the ability of network.
		\section{Related work}
		
		In today's society, writer identification is playing an increasingly important role in the fields of finance, insurance, public security departments, criminal investigations, and court trials\cite{8237850,6117473,GUERBAI2015103,HAFEMANN2017163}. According to the type of handwriting, writer identification can be divided into text-related writer identification and text-independent writer identification; according to approaches used for writer identification, writer identification can be divided into traditional handcrafted approaches and deep learning approaches.
		
		Text-related writer identification is that the contents of the handwriting to be recognized are the same. The typical text-related writer identification is signature verification. Many writer identification work based on this\cite{8237850,6117473,HAFEMANN2017163,8100166}. Text-independent writer identification is that the content of the handwriting to be recognized is different. Since text-independent writer identification does not rely on text, it has wider applicability than text-related writer identification, and it is also more difficult to recognize. The global regular network(GRN) proposed in this paper is text-independent writer identification.
		
		Handcrafted approaches for writer identification were based on feature extraction. Many feature extraction algorithms, such as Gabor filter\cite{HELLI20102199}, Local binary patterns (LBP)\cite{BERTOLINI20132069,YILMAZ2016109}, and SIFT\cite{CHRISTLEIN2017258,8270096,8395190} are applied to feature extraction. Then various classifiers are used to classify, such as distance-based classifier, Support Vector Machines
		(SVM)\cite{1030921}, Hidden Markov Model (HMM)\cite{SAID2000149}, Fuzzy based classifier\cite{TAN20093313} and so on. These methods have laid a solid foundation for writer identification. Handcrafted approaches have achieved good results on some datasets, but it is difficult to apply to a wider range of handwriting identification. In order to overcome the difficulties faced by feature point extraction algorithms, convolutional neural networks have been applied to writer identification\cite {10.1007/978-3-319-23117-4_3,10.1007/978-3-319-24947-6_45,7814125,7814128,NGUYEN2019104,9040654}. One of the first attempts at writer identification with CNN was proposed by Fiel and Sablatnig\cite{10.1007/978-3-319-23117-4_3} in 2015. They used CNN to extract features and the feature vectors output from CNN will be compared with  precalculated feature vectors of
		the dataset by nearest neighbor classification. Christlein et al.\cite{10.1007/978-3-319-24947-6_45} proposed a similar network. The different place is that they encoded the feature vectors with Gaussian mixture models (GMM) super
		vector encoding. Above these, Xing et al.\cite{7814128} proposed a method based on pixel scanning strategy, data augmentation and combined multi-stream parallel CNN. On the standard dataset of HWDB\cite{6065272}, obtaining 97.0\% recognition rate using small amount of handwriting. However, these networks only use single character or smaller pixel blocks as input to extract the local features, ignoring the handwriting features contained in the larger handwriting scale and failing to extract more handwriting features.
		
		The attention mechanism\cite{WOS:000181471800047,Itti2001,WOS:000452647102103,7807286} is an effective way to capture stroke information. Chen et al.\cite{10.1007/978-3-030-01240-3_15} used the attention reversal mechanism to capture prominent objects. Inspired by the attention mechanism, Wang et al.\cite{8100166} proposed the residual attention network. This kind of network has the following appealing properties: it can incorporate with state-of-art feed forward network architecture in an end-to-end training fashion; attention modules can extract features in different levels. In experiment, this method achieves 0.6\% top-1 accuracy improvement with 46\% trunk depth and 69\% forward Flops comparing to ResNet-200. In our work, residual attention network is used for local feature extraction.
		\section{Global Regular Network}
		
		Different people have different handwriting, the characteristics of these handwriting are not only contained in the local characteristics of the handwriting, such as the thickness, length, inclination degree, degree of curvature of each stroke, and distribution of strokes, but also in the global characteristics of the handwriting, such as the distribution of words, the arrangement of the rows and columns in the handwriting. These global features of handwriting contain a wealth of information. The traditional handwriting feature extraction method based on the handwriting pixel block can often only capture the local features of the handwriting, but cannot effectively capture the global information, which is a great waste of dataset resources. 
		
		The global regular network(GRN) proposed in this paper captures both the global and local characteristics of handwriting, realizing the adequate exploitation of handwriting.
		\begin{figure}[!t]
			\begin{center}
				%\fbox{\rule{0pt}{2in} \rule{0.9\linewidth}{0pt}}
				%#\includegraphics{image1.eps }
				%\includegraphics[width=0.8\linewidth]{image1.eps}
				\includegraphics[width=2.5in]{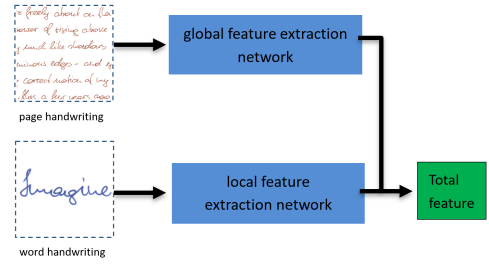}
			\end{center}
			\caption{Illustration of Global Regular Network}
			\label{image1}
		\end{figure}
	\begin{figure*}
		\begin{center}
			%\fbox{\rule{0pt}{2in} \rule{.9\linewidth}{0pt}}
			\includegraphics[width=1\linewidth]{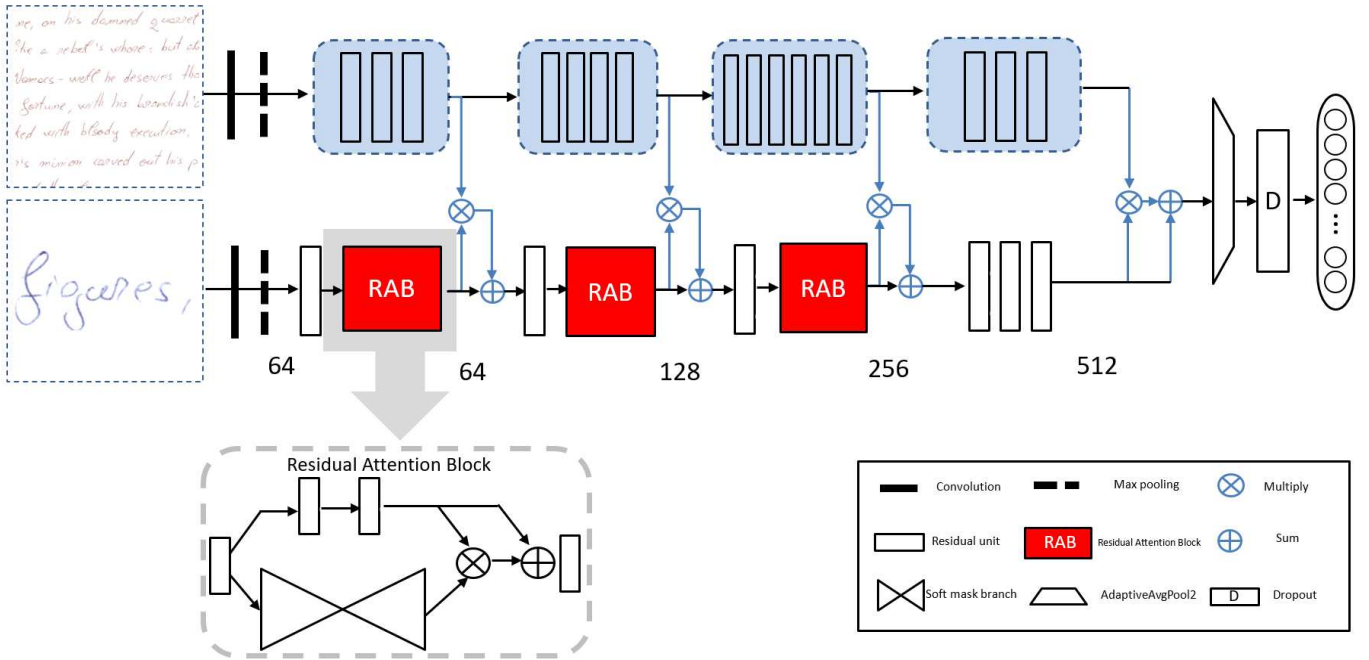}
		\end{center}
		\caption{The overall architecture of Global Regular Network}
		\label{image2}
	\end{figure*}
		We illustrate our thoughts with Fig.\ref{image1}. The network extracts the handwriting features in two branches, one extracts the global features of the handwriting, another one extracts the local features of the handwriting, and then combines the two features as the general features of the handwriting. For input images, We believe that to extract the global characteristics of the handwriting, the input data must be the page scale of the handwriting including multiple words, so that the handwriting will contain rich global features; and to extract the local features of the handwriting, the single word handwriting meet the requirement, the reason is: too large region of handwriting will import global features (such as the distribution of different words), which will become the noise of local feature extraction; too small region of handwriting such as a single stroke will lose many local features (such as continuous writing in word handwriting), so single word handwriting is the ideal region for local feature extraction. For the merge of local features and global features, suppose page handwriting and word handwriting are x and y, respectively. We get global handwriting feature G(x) after page handwriting going through the global feature extract network and local handwriting feature L(y) after word handwriting going through the local feature extract network. The general feature T(x, y) is obtained like below:
		\begin{equation}  
			T(x,y)=(1+G(x))\times L(y)
			\label{equ1}
		\end{equation}
		In this formula, the global features of the handwriting act as weight parameters to adjust the local features. We use the combination of local features and global features in Equ.\ref{equ1} instead of the combination like $T(x, y)=G(x)\times L(y)$, because we believe that the local features of the handwriting contain the main handwriting information, and should be taken as the main body to characterize the handwriting features while global features just are taken as parameters to adjust the local handwriting features, playing a role in the global regularization of the local features, which is also the origin of our network name. In addition, in Equ.\ref{equ1}, even if the global feature is 0, the network will still work correctly, and the network result after adding global feature extraction network will only be better. In Sec.\ref{sec4-3}, we show the effects of different combinations of global and local features on the experimental results.
		\subsection{Architecture}
		
		In Fig.\ref{image2}, the input handwriting consist of the above page handwriting and below word handwriting. The global regular network consists of the global feature extraction network above and the local feature extraction network below.
		Christlein et al.\cite{10.1007/978-3-319-24947-6_45} verify that resnet\cite{10.1007/978-3-319-46493-0_38} networks can extract handwriting better than other networks. Considering GPU occupancy, parameters number, and training efficiency, we chose resnet34 as the overall architecture of the global feature extraction network which include: a convolution layer, initializing the convolution kernel to 64, a $3\times 3$ maximum pooling layer, and 4 residual modules. The number of residual units in each residual module is 3, 4, 6, 3, respectively. The final output features merge with local feature, forming general feature. Compared with the classic resnet34 model, the global feature extraction network has been revised in two parts: the convolutional layer convolutional kernel size of the input image increased from 3 to 7, because we believe that the larger convolutional kernel will be more conducive to pay attention to the global characteristics of the handwriting; the second change is to insert the maximum pooling layer after the first convolutional layer, in order to match global feature with local feature.
		
        Residual attention network proposed by Wang et al.\cite{8100166} is used to extract local features. As shown in the below branch of Fig.\ref{image2}, residual attention network includes: a convolutional layer to initialize the convolution kernel to 64, a $3\times 3$ maximum pooling layer, and a series of residual attention blocks and residual units. Details of residual attention network have been listed in Table.\ref{table1}. We also can see the structure of residual attention blocks: the trunk branch containing two residual units and another soft mask branch using bottom-up top-down structure\cite{7478072,10.1007/978-3-319-46484-8_29,7410535}. The outputs of trunk branch and soft mask branch merge with a residual way. Detail of residual attention blocks are showed in Sec.\ref{sec3-2}.
		
		Four global residual connections were used between global feature extraction network and local feature extraction network. The purpose of these connections is to make the global features adequately work on the local features. Finally, the general handwriting features flow into the average pooling layer obtaining $1\times 1$ handwriting features; due to the small dataset, we add a dropout layer to the network to reduce the over-fitting. Full connected layer is connected to the end. Detail of overall architecture is showed in Table.\ref{table1}.
		
		The entire network is trained in end-to-end fashion. Two mechanisms are used for writer identification. First, two branches are used to extract richer characterizes in handwriting. The global features of the handwriting act as weight parameters to adjust the local features, playing a role in the global regularization of the local features. The second is that residual attention module is used to extract the local features of the handwriting.
		\subsection{Residual Attention Block}
		\label{sec3-2}
		\begin{figure}[!t]
			\begin{center}
				%\fbox{\rule{0pt}{2in} \rule{0.9\linewidth}{0pt}}
				%#\includegraphics{image1.eps }
				\includegraphics[width=2.5in]{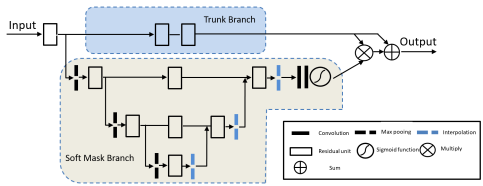}
			\end{center}
			\caption{The architecture of Residual Attention Block}
			\label{image3}
		\end{figure}
		Residual attention blocks are used to extract local features. The architectures are showed in \ref{image3}: the branch above is trunk branch, which is the main path of network; another below is soft mask branch. Soft mask branch uses bottom-up top-down structure\cite{7478072,10.1007/978-3-319-46484-8_29,7410535}. Input features go through three maximum pooling layers and reach minimum resolution; Then bilinear interpolation layers with the same number of maximum pooling layers operate on the intermediate features. The size of output features is same with the input features. In order to extract more information contained in input with different size, two another branches were added in where the input size reduce two times and four times. Then the output go through two convolutional layers and a sigmoid normalization layer which scales the value of output between(0, 1). Finally, similar to the combination of global features and local features, the features of two branches also carry out the same strategy: Suppose the input feature is x, the output feature of trunk branch is T(X), the output of soft mask branch is M(X), and the final output is:$H(X)=(1+ M(X))\times T(X)$.
		\begin{table*}[t]
			\caption{GLOBAL REGULAR NETWORK ARCHITECTURE DETAILS}
			\label{table1}
			\centering
			\begin{tabular}{c|c|c|c|c}
				\hline
				\textbf{Layer-Res} & \textbf{Layer—Att} & \textbf{Resnet}&\textbf{Attention}&\textbf{Output Size}\\
				\hline
				\multicolumn{2}{c|}{Conv1} &\multicolumn{2}{c|}{$7\times7$,64,stride 2} & $128\times128$ \\
				\hline
				\multicolumn{2}{c|}{Max pooling}  &\multicolumn{2}{c|}{$3\times3$,stride 2} & $64\times64$ \\
				\hline
				\multirow{3}{*}{Residual Unit}&\multirow{2}{*}{Residual Unit}& \multirow{3}{*}{$ \left(\begin{array}{c}
						1\times1,64\\	
						3\times3,64\\
					\end{array}
					\right)\times 3
					$}&\multirow{2}{*}{$ \left(\begin{array}{c}
						1\times1,64\\	
						3\times3,64\\
					\end{array}
					\right)\times 1
					$}&\multirow{3}{*}{$64\times64$}\\
				\multirow{3}{*}{} & & & & \\
				\cline{2-2}\cline{4-4}
				\multirow{3}{*}{} & Attention Module & \multirow{3}{*}{} & $Attention\times1$ &  \\
				\hline
				\multirow{3}{*}{Residual Unit}&\multirow{2}{*}{Residual Unit}& \multirow{3}{*}{$ \left(\begin{array}{c}
						1\times1,128\\	
						3\times3,128\\
					\end{array}
					\right)\times 4
					$}&\multirow{2}{*}{$ \left(\begin{array}{c}
						1\times1,128\\	
						3\times3,128\\
					\end{array}
					\right)\times 1
					$}&\multirow{3}{*}{$32\times32$}\\
				\multirow{3}{*}{} & & & & \\
				\cline{2-2}\cline{4-4}
				\multirow{3}{*}{} & Attention Module & \multirow{3}{*}{} & $Attention\times1$ &  \\
				\hline
				\multirow{3}{*}{Residual Unit}&\multirow{2}{*}{Residual Unit}& \multirow{3}{*}{$ \left(\begin{array}{c}
						1\times1,256\\	
						3\times3,256\\
					\end{array}
					\right)\times 6
					$}&\multirow{2}{*}{$ \left(\begin{array}{c}
						1\times1,256\\	
						3\times3,256\\
					\end{array}
					\right)\times 1
					$}&\multirow{3}{*}{$16\times16$}\\
				\multirow{3}{*}{} & & & & \\
				\cline{2-2}\cline{4-4}
				\multirow{3}{*}{} & Attention Module & \multirow{3}{*}{} & $Attention\times1$ &  \\
				\hline
				\multirow{3}{*}{Residual Unit}&\multirow{3}{*}{Residual Unit}& \multirow{3}{*}{$ \left(\begin{array}{c}
						1\times1,512\\	
						3\times3,512\\
					\end{array}
					\right)\times 3
					$}&\multirow{3}{*}{$ \left(\begin{array}{c}
						1\times1,512\\	
						3\times3,512\\
					\end{array}
					\right)\times 3
					$}&\multirow{3}{*}{$8\times8$}\\
				\multirow{3}{*}{} & & & & \\
				
				\multirow{3}{*}{} &  & \multirow{3}{*}{} &  &  \\
				\hline
				\multicolumn{2}{c|}{Average pooling}  &\multicolumn{2}{c|}{$8\times8$,stride 1} &$1\times1$ \\
				\hline
				\multicolumn{2}{c|}{Dropout,FC}  &\multicolumn{3}{c}{1000} \\
				\hline
			\end{tabular}
			
		\end{table*}
		\section{Result}
		In this chapter, we introduce our experiment in three parts. The first part introduces the dataset used in training, data preprocessing and the settings used in training. In the second part, we show the results of our network, which exceeds most state-of-the-art writer identification structures. In the third part, we show the effectiveness of the two highlights in this article: the effectiveness of using global regularization strategies and the effectiveness of using residual attention network.
		
		\subsection{Data Preprocessing and Training Settings}
		The dataset of the CVL\cite{6628682} contains 27 writers in training dataset and 283 writers in test dataset. Every writer contributes 7 writing materials and 5 writing materials in the training dataset and test dataset respectively. Writers contribute a German material, and the rest are English materials. Each piece of material consists of three parts: page handwriting, line handwriting segmented from page handwriting, and word handwriting segmented from page handwriting.
		
		We preprocess the dataset. We merge the original training dataset and test dataset to form a new dataset. It contains 310 writers, 1550 page handwriting, and 96180 word handwriting. Since the page handwriting are not enough, we do data augmentation on this part. We randomly divide one page handwriting into 9 smaller square handwriting as new page handwriting to increase the number of page handwriting and cut off the blank areas around the handwriting before segmentation to avoid the import of noise. After that, the new page handwriting is scaled down to $256\times 256$ size. For word handwriting, because of its different sizes, we first fill the picture to square shape and then reshape to $256\times 256$ size.
		During training, each batch consists of one word handwriting and a random page handwriting from the same writer. It should be noted that the word handwriting and the page handwriting are from the same writer, but not necessarily from the same written material. Since even the same writer, the handwriting written in different environments will change, these pairs of input handwriting will force the network to capture the same characteristics of the handwriting written in different environments. We train the model based on pytorch1.7.1 platform
		with PTX-P2 GPU. The optimizer uses Adam, the initial learning rate is 0.001, and half every 30 epochs.
		
		\subsection{Comparison With State of Art Structures}
		\label{sec4-2}
		We compare the training results with several other structures. Fiel et al.\cite{10.1007/978-3-319-23117-4_3} firstly used CNN for writer identification. CNN was used to extract features, and feature vectors output from CNN will be compared with  precalculated feature vectors of
		the dataset by nearest neighbor classification. Christlein et al.\cite{10.1007/978-3-319-24947-6_45} proposed a similar network. The different place is that they encoded the feature vectors with Gaussian mixture models (GMM) super
		vector encoding. WU et al.\cite{6716030} uses an isotropic log filter to divide the handwriting into word sizes, and then extracts the SIFT descriptor of the handwriting and then matches the registered handwriting. DLS-CNN\cite{Chen2020} first performs line segmentation on the page handwriting, then cuts each line into small pixel blocks of the same size, and uses resnet50 for writer identification at the pixel block level. Table.\ref{table2} shows that our structure has achieved the best performance with results of 99.98\% top1 accuracy and 100\% top5 accuracy. We attribute it to the combination of global and local features. The previous network focused on the extraction of local features, and our network added a branch to extract global features, allowing the network to extract more handwriting features. Compared with the DLS-CNN network, we used a smaller model(resnet34) but achieved better results, which reflects the effectiveness of our strategy. In the experimental results, the Top-1 of DLS-CNN are much lower than other networks. We attribute it as the input image pixel block is too small (such as the stroke level, smaller than the word size), such handwriting will lose many word level information (distribution of strokes and continuous strokes) which greatly reduce the training accuracy of the network.
		
		\begin{table}[!t]
			\caption{COMPARISON OF GRN AND OTHER PROPOSED WRITER IDENTIFICATION STRUCTURES}
			\label{table2}
			\begin{center}
				\begin{tabular}{lccr}
					\hline
					& Input materials & Top-1 & Top-5 \\
					\hline
					Fiel\cite{10.1007/978-3-319-23117-4_3} & one page & 98.9 & 99.3 \\
					Wu\cite{6716030} & one page & 99.2 & 99.5 \\
					Christlein\cite{10.1007/978-3-319-24947-6_45} & one page & 99.4 & N/A\\
					Tang\cite{7814125} & one page & 99.7 & 99.8 \\
					DLS-CNN\cite{Chen2020} & 256 pixel & 95.8& 99.9\\
					Ours & page+word &\pmb{99.92} & \pmb{100}\\
					\hline
				\end{tabular}
			\end{center}
		\end{table}
		
		\subsection{Comparison With Different Structures}
		\label{sec4-3}
		\begin{figure}[!t]
			\begin{center}
				%\fbox{\rule{0pt}{2in} \rule{0.9\linewidth}{0pt}}
				%#\includegraphics{image1.eps }
				\includegraphics[width=2.5in]{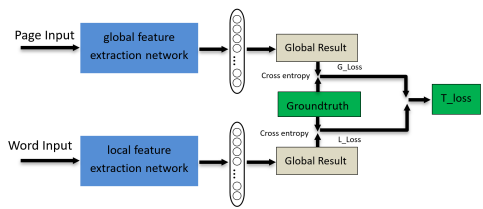}
			\end{center}
			\caption{The structure of Net1}
			\label{image4}
		\end{figure}
		We have designed two structures of the network to compare with our global regular network. One structure is shown in Fig.\ref{image4}. We mark it Net-1. The global feature extraction network and the local feature network in this structure are not connected. The page handwriting go through the above global feature extraction network, the output calculate the cross entropy loss with corresponding ground truth, we mark it Loss-G; similarly, word handwriting go through below local feature extraction network, the output calculate the cross entropy loss with corresponding ground truth, we mark it Loss-L. The total loss is a convex combination of Loss-G and Loss-L:
		$Loss-T = k\times Loss-G+(1-k)\times Loss-L $
		where k is a hyperparameter. The region of k is (0-1). 0,0.5,1 are assigned to k for training respectively. When k=0, the network only updates Loss-L, which mains just the local feature extraction network takes part in training; when k=1, the network only updates Loss-G, which mains just the global feature extraction network takes part in training; when k=0.5, Both the global feature network and the local feature network will be added to the training, and this is another combination method that is different from the global residual connection.
		
		The second structure is almost same to global regular network, but just saving the last global residual connection. We mark it Net2. The comparison of these structures with GRN will verify the effect of the adequacy of global residual connection.
		\begin{table}[!t]
			\caption{COMPARISON OF GRN AND OTHER DESIGNED STRUCTURES}
			\label{table3}
			\begin{center}
				\begin{tabular}{lccr}
					\hline
					Handwriting & Train-Loss & Test-loss \\
					\hline
					Net1-word  & 0.010 & 0.052 \\
					Net1-page  & 0.031 & 0.057\\
					Net1-both  & 0.020& 0.040\\
					Net2  & 0.0074& 0.008\\
					Ours  &\pmb{0.0065} & \pmb{0.004}\\
					\hline
				\end{tabular}
			\end{center}
		\end{table}
	
		For the consideration of training efficiency, the dataset used for the comparison experiment just need to be the same, we randomly select 60 classes from the original dataset of 310 classifications as the new dataset for the comparison experiment. In addition, due to the small number of datasets, we found that the accuracy of some networks easily reach 100\% on some iterations in the final test, and the accuracy index will no longer effectively reflect the ability of the network. And we found that the loss index will remain at a stable level. So we choose the loss in training process and testing process as the evaluation index of network writer identification ability. The lower the train-loss and test-loss, the better the result. The experimental results are shown in Table\ref{table3}.
		
		First, we compare the three experiments of Net1. For train-loss, Net1-page is 0.031, Net1-both is 0.020, and Net1-word is 0.010. We have two findings: one is that the training effect of using page as dataset is not well as using word; the other is, the smaller rate of loss-G in loss-T, the higher the accuracy of training. We conclude that the training effect of page handwriting is not as good as that of word handwriting. We attribute it to the fact that human handwriting features are mainly contained in local features. For test-loss, Net1-page is 0.057, Net1-both is 0.040, and Net1-word is 0.052. We found that the effect of word input is better than that of page input, but worse than the test result obtained by mixed input of word and page. We attribute it to the fact that the overfitting introduced by using only one type of handwriting as input and the mixed input allows the network to classify the handwriting from multiple sources, resulting in the introduction of regularization to reduce the effect of overfitting. The second network train-loss is 0.0074, test-loss is 0.008, and the effect is much better than the first three networks. This reflects the effectiveness of global residual connection methods. Compared with Net2, our network processes global residual connections on the four modules, and achieves train-loss and test-loss of 0.0065 and 0.004. The performance is further improved on the basis of Net2, which reflects the validity of the four-way global residual connection model. The greatest improvement is test-loss, which we attribute to the four-way global residual connection that makes the local features and global features more fully merged, and the regular effect of global features is further improved.
		
		We also verify the efficiency of the residual attention network to extract local features. We design a comparative experiment: only use the traditional resnet34 network to extract local features. That means the global feature extraction network and the local feature extraction network have the same structure. The result compared with attention residual network is shown in Table.\ref{table4}. The results show that there is a big gap between the network performance without the attention module and with the attention module. On the other hand, in actual experiments, the network using the attention module can be trained at a faster speed.
		\begin{table}[!t]
			\caption{COMPARISON OF GRN WITH RESIDUAL ATTENTION NETWORK AND GRN WITHOUT RESIDUAL ATTENTION NETWORK}
			\label{table4}
			\begin{center}
				\begin{tabular}{lccr}
					\hline
					Handwriting & Train-Loss & Test-loss \\
					\hline
					Net-without attention  & 0.025 & 0.023 \\
					Net-with attention  & \pmb{0.0065} & \pmb{0.004}\\
					\hline
				\end{tabular}
			\end{center}
		\end{table}
		
		\section{Conclusion}
		We propose global regular network (GRN) for writer identification. The network has two highlights: First, the network pays attention to the global characteristics of handwriting. The network extracts handwriting features in two ways: one is to extract the global features of the handwriting, and the other is to extract the local features of the handwriting. Finally, the global feature and the local feature are combined in a global residual way. With this operation, more information will be extracted by our network; the second is that GRN uses the residual attention network to extract the local features of the handwriting, so that the network can extract more local features. We performed writer identification tasks on CVL dataset, and achieved amazing results of 99.98\% top1 accuracy and 100\% top5 accuracy, surpassing the current state-of-the-art architecture. The core idea of ​​the network design is to extract the features in multiple levels. We can not only use this strategy in the field of writer identification, but also applies to many other areas of feature extraction.
%\IEEEtriggeratref{8}
% The "triggered" command can be changed if desired:
%\IEEEtriggercmd{\enlargethispage{-5in}}
\bibliographystyle{IEEEtran}
\bibliography{GRN}
	\end{document}